\documentclass[sigconf,screen]{acmart}

\usepackage[linesnumbered,ruled,vlined]{algorithm2e}
\usepackage{float}
\usepackage{tabularx}
\usepackage{enumitem}
\usepackage{subcaption}
\usepackage{booktabs} %

\copyrightyear{2026}
\acmYear{2026}
\setcopyright{cc}
\setcctype{by}
\acmConference[LAK 2026]{LAK26: 16th International Learning Analytics and Knowledge Conference}{April 27-May 01, 2026}{Bergen, Norway}
\acmBooktitle{LAK26: 16th International Learning Analytics and Knowledge Conference (LAK 2026), April 27-May 01, 2026, Bergen, Norway}
\acmDOI{10.1145/xxxxxxx}
\acmISBN{979-8-xxxxxx}

\begin{document}

\title[Reasoning Trace Analytics in Multi-Agent Systems]{Disagreement as Data: Reasoning Trace Analytics in Multi-Agent Systems}

\author{Elham Tajik}
\orcid{}
\affiliation{
  \institution{University at Albany}
  \city{Albany, NY}
  \country{USA}
}
\email{etajik@albany.edu}

\author{Conrad Borchers}
\authornote{Both authors contributed equally to this research.}
\orcid{0000-0003-3437-8979}
\affiliation{
  \institution{Carnegie Mellon University}
  \city{Pittsburgh, PA}
  \country{USA}
}
\email{cborcher@cs.cmu.edu}

\author{Bahar Shahrokhian}
\authornotemark[1]
\orcid{0000-0002-3737-4714}
\affiliation{
  \institution{Arizona State University}
  \city{Tempe, AZ}
  \country{USA}
}
\email{bshahrok@asu.edu}

\author{Sebastian Simon}
\orcid{0000-0003-3218-2032}
\affiliation{
  \institution{Le Mans University}
  \city{Le Mans, Pays de la Loire}
  \country{France}
}
\email{sebastian.simon@univ-lemans.fr}

\author{Ali Keramati}
\orcid{0000-0002-4763-7418}
\affiliation{
  \institution{University of California, Irvine}
  \city{Irvine, CA}
  \country{USA}
}
\email{a.kera@uci.edu}

\author{Sonika Pal}
\orcid{0009-0007-3470-6369}
\affiliation{
  \institution{Indian Institute of Technology Bombay}
  \city{Mumbai, Maharashtra}
  \country{India}
}
\email{sonika_pal@iitb.ac.in}

\author{Sreecharan Sankaranarayanan}
\orcid{0000-0001-9122-6870}
\affiliation{
  \institution{Extuitive Inc. (Flagship Pioneering)}
  \city{Cambridge, MA}
  \country{USA}
}
\email{sreecharan.primary@gmail.com}

\renewcommand{\shortauthors}{Tajik et al.}

\begin{abstract}
Learning analytics researchers often analyze qualitative student data such as coded annotations or interview transcripts to understand learning processes. With the rise of generative AI, fully automated and human--AI workflows have emerged as promising methods for analysis. However, methodological standards to guide such workflows remain limited. In this study, we propose that reasoning traces generated by large language model (LLM) agents, especially within multi-agent systems, constitute a novel and rich form of process data to enhance interpretive practices in qualitative coding. We apply cosine similarity to LLM reasoning traces to systematically detect, quantify, and interpret disagreements among agents, reframing disagreement as a meaningful analytic signal. Analyzing nearly 10,000 instances of agent pairs coding human tutoring dialog segments, we show that LLM agents' semantic reasoning similarity robustly differentiates consensus from disagreement and correlates with human coding reliability. Qualitative analysis guided by this metric reveals nuanced instructional sub-functions within codes and opportunities for conceptual codebook refinement. By integrating quantitative similarity metrics with qualitative review, our method bears potential to improve and accelerate the process of establishing inter-rater reliability during coding by surfacing interpretive ambiguity, especially when LLMs collaborate with humans. We discuss how reasoning-trace disagreements represent a valuable new class of analytic signals advancing methodological rigor and interpretive depth in educational research.
\end{abstract}



\maketitle

\section{Introduction}
The field of learning analytics has increasingly turned to artificial intelligence (AI) to interpret and make sense of learner data. With the rise of large language models (LLMs), our field's initial focus on log data artifacts to understand learning \cite{mohammadi2025artificial} has increasingly shifted to textual data. In particular, learning analytics researchers often interpret complex textual data produced by students and other stakeholders, such as interview data \cite{borchers2025learner,keramati2024students}, tutoring transcripts \cite{thomas2025does}, think-aloud data \cite{borchers2025large}, and open-response data \cite{botelho2023leveraging}. Many of these data types were traditionally not analyzable through AI and largely confined to small-scale human analysis. Therefore, to date, shared methodological standards for humans to co-analyze these data with AI are lacking, with very few empirical studies, for instance, on the human--AI co-creation of qualitative codebooks \cite{barany2024chatgpt}.

Toward creating said standards, we ground this study in a particularly common form of qualitative data analysis in our field, that is, qualitative coding, where humans annotate segments of data. Qualitative coding has been widely used in learning analytics, for instance, in developing detectors \cite{baker2024detector} or analyzing interview transcripts \cite{borchers2025learner}. Despite the advances of using AI for this task, prior work emphasizes the continued need for human oversight to ensure reliability, interpretability, and accuracy \cite{de2024performing, barany2024chatgpt, liu2025qualitative}. What this oversight should look like, and how we can use LLMs to identify cases worth allocating human attention to, are both open questions of particular importance as our field moves toward analyzing data with LLMs at an increasingly large scale.

We propose a novel approach: treating reasoning traces produced by LLMs employed for qualitative coding as a new form of process analytics. Reasoning traces are the intermediate steps and justifications that agents generate when arriving at a decision, which make visible not only outcomes but also the learned semantic and statistical regularities LLMs activate when generating responses to particular prompts. However, as recent studies such as Kambhampati et al. \cite{kambhampati2025stop} have cautioned, these traces should not be mistaken for genuine human-like reasoning. Anthropomorphizing such outputs risks overstating the epistemic transparency of LLMs and ignoring their fundamentally probabilistic, pattern-matching nature. To address this concern, we treat reasoning traces not as evidence of authentic cognition but as structured textual artifacts that can be systematically analyzed to probe the reliability and conceptual accuracy of LLMs during coding. In particular, rather than discarding agent reasoning traces, we introduce a method to detect and interpret them as productive sites of analysis. Our approach leverages sentence embeddings, which encode textual segments into high-dimensional vectors such that semantically similar sentences are positioned closer together. By quantifying the similarity of agent reasoning through cosine distance between these embeddings, we can systematically identify points of conceptual divergence. Combining this quantitative signal with qualitative inspection, we show how disagreement cases can be surfaced for human review, opening pathways toward more efficient and reliable coding, richer interpretive insights, and human--AI collaborative workflows in learning analytics. We ask:

\begin{enumerate}[label=\textbf{RQ\arabic*:}, leftmargin=*]
    \item How can disagreements in agent reasoning be systematically detected and represented as analytic signals?
    \item What patterns in agent disagreement reasoning reveal hidden uncertainty in qualitative analysis?
\end{enumerate}

\section{Background and Motivation}

\subsection{LLMs in Learning Analytics}
Digital learning environments generate large, complex datasets ranging from eye tracking and clickstreams to classroom video and think-aloud protocols. These data provide rich insights into student cognition and behavior \cite{borchers2025large}. As data complexity grows, learning analytics is shifting beyond purely quantitative approaches toward qualitative methods that capture the multifaceted nature of learning \cite{bakharia2025transcripts}. Within this domain, process data, i.e., detailed records of learner interactions and cognitive activity, has emerged as a focal point for educational data mining and AI research \cite{mazzullo2023learning}. One widely used qualitative method, Thematic Analysis (TA), enables detection and interpretation of patterns in datasets, either inductively or guided by predefined theories or codebooks \cite{braun2006using,braun2021one,liu2025qualitative}. However, manual thematic analysis is time- and resource-intensive \cite{braun2006using}. This challenge has spurred growing interest in leveraging LLMs to streamline and scale qualitative analyses in learning analytics.

\subsection{LLMs for Qualitative Analysis}
Traditional machine learning techniques, such as Naïve Bayes classifiers and clustering, have offered modest automation within qualitative workflows but still require extensive human interpretation and are limited in scalability \cite{sherin2018learning}. LLMs, with their advanced natural language understanding and generation capabilities, present new possibilities for qualitative coding and analysis \cite{zambrano2023ncoder,yan2024human}. For example, Ramanathan et al., \cite{ramanathan2025prompt}, applied the GROPROE iterative prompt refinement framework to improve deductive coding of student reflections. Nevertheless, these single-agent LLM approaches remain limited to the LLM's context window (number of tokens an LLM can process at a time) and the performance of attention mechanisms (responsible for selecting relevant tokens), thus constraining the amount of text they can process in a single interaction, and limiting their ability to analyze large datasets holistically without human oversight \cite{barany2024chatgpt}. To address this, researchers have increasingly adopted Multi-Agent Systems (MAS) approaches, distributing coding responsibilities across specialized roles such as coders, reviewers, and adjudicators, that engage in iterative dialogue and consensus-building \cite{sankaranarayanan2025automating,simon2025comparing,qiao2025thematic,rasheed2024can}. Borchers et al., \cite{borchers2025temperature}, examined the effects of parameters such as temperature and persona, finding that these adjustments only rarely improve results, thus suggesting that the key advantage of MAS does not lie in parameter tuning but in the collaborative structure itself. This positions MAS as a promising paradigm for robust qualitative coding.

\subsection{Evaluating Text Similarity with LLMs}
In natural language processing (NLP), text similarity can be assessed using lexical overlap metrics such as BLEU \cite{papineni2002bleu}, ROUGE \cite{lin2004rouge}, and METEOR \cite{banerjee2005meteor}, or semantic metrics like cosine similarity. Unlike lexical metrics, cosine similarity measures the angle between vector embeddings of text, effectively capturing semantic relatedness independent of length \cite{gomaa2013survey,schutze2008introduction}. Advances in distributed word embeddings \cite{mikolov2013efficient} and contextualized models like BERT \cite{devlin2019bert} have popularized cosine similarity for comparing text at multiple granularities, including utterances and documents. Educational applications include automated essay scoring, where cosine similarity is leveraged to compare student responses and expert references \cite{lahitani2016cosine}. Despite this, semantic similarity methods remain underutilized in learning analytics. In this work, this paper proposes using reasoning traces from large language model agents as a new form of process data for learning analytics. By measuring similarity between agents' reasoning, the study shows that disagreements can be systematically detected and meaningfully interpreted. The approach improves coding reliability, reveals ambiguities in codebooks, and supports effective human--AI collaborative analysis in educational research.

\section{Methods}

\subsection{Dataset}
The dataset for this study originates from Barany et al. \cite{barany2024chatgpt}, which investigated high-dosage mathematics tutoring involving trained tutors delivering personalized, small-group instruction. Transcripts come from 60-minute virtual tutoring sessions with 9th-grade Algebra I students attending high-poverty urban schools in the northeastern United States during 2022 and 2023. For this analysis, we extracted 3,538 tutor--student dialogue segments sampled from three distinct tutoring sessions, treating each segment as an independent data point.

\subsection{Codebook Development}
\label{sec:method:codebook}
We adopted a hybrid human--AI codebook originally developed by Barany et al. \cite{barany2024chatgpt}, which demonstrated superior coding reliability and interpretability compared to fully LLM-generated codebooks.\footnote{\url{https://github.com/bshahrok/Disagreement-as-Data-LAK-2026}} This codebook was built through iterative refinement: three tutoring transcripts were segmented into 13 smaller batches for initial coding, while a fourth transcript comprising 150 lines was used to refine code definitions. Each utterance was double-coded by two trained annotators using binary labels, establishing the ground truth across eight categories (see also \cite{barany2024chatgpt}):

\begin{description}[leftmargin=0pt, labelsep=1em]
    \item[\textbf{Greeting:}] Salutations or farewells at the start or end of a session (e.g., ``Hello,'' ``Enjoy the rest of your day'').
    \item[\textbf{Instruction:}] Directives related to lesson tasks (e.g., ``Go ahead and fill that out'').
    \item[\textbf{Guiding Feedback:}] Clarifying or scaffolding responses to student work (e.g., ``Not quite. Look for factor pairs'').
    \item[\textbf{Aligning to Prior Knowledge:}] References to previously learned concepts, often featuring the word ``remember'' (e.g., ``Remember Giovanni, what does factor mean?'').
    \item[\textbf{Understanding/Engagement-Tutor:}] Tutor questions checking for understanding (e.g., ``Why do you think we might have done that?'').
    \item[\textbf{Technical or Logistics:}] Comments about technical issues like audio or video quality (e.g., ``You're on mute,'' ``Can you hear me okay?'').
    \item[\textbf{Encouragement:}] Positive affirmations of student effort or accuracy (e.g., ``Good job,'' ``You're getting it, man'').
    \item[\textbf{Time Management:}] Statements concerning timing or pacing (e.g., ``We have about 5 minutes left'').
\end{description}

\subsection{Multi-Agent System}
Our MAS builds on the system developed by Borchers et al. \cite{borchers2025temperature}, designed for AI-assisted deductive coding using a predefined codebook. It has four components: (1) a Dual-Agent Discussion module, (2) a Consensus Agent, and (3) a hybrid human--AI benchmark dataset. All agents employ the same codebook for consistency. Agent roles are as follows:
\textit{\textbf{Dual-Agent Discussion:}} Two agents with distinct personalities (e.g., bold versus empathetic) independently generate codes based on the codebook during Round 1. If consensus is not reached, the agents engage in a Round 2 dialogue wherein they critique and revise each other's codes to reconcile differences.
\textit{\textbf{Consensus Agent:}} A neutral, balanced agent reviews the outputs and reasoning of the two discussion agents to resolve disagreements and produce a final consensus code aligned with the codebook. Even where partial agreement exists, the consensus agent ensures completeness and consistency in the final coding decision.

\subsubsection*{\textbf{Model Selection}}
Building on prior evaluations of multiple open-source LLMs \cite{borchers2025temperature}, we selected DeepSeek-R1-32B \cite{guo2025deepseek} as the primary model for this study. This choice was motivated by the following considerations. First, DeepSeek-R1 outputs results alongside reasoning traces \cite{kambhampati2025reasoning}, rather than only final answers, making it well-suited for our focus on interpretive alignment. Second, the DeepSeek-R1 models exemplify a trend toward optimization of output quality per used parameter: While rivaling models with up to 175 billion parameters (e.g., OpenAI-o1-1217), DeepSeek-R1 effectively only activates 37 billions per token, optimizing computational costs. The 32B dense variant approximates the performance of the bigger DeepSeek-R model in key reasoning tasks and benchmarks such as AIME2024, further improving deployability and computation costs. Such considerations are essential for operating Multi-Agent systems with a high number of expected calls. Overall, leveraging DeepSeek-R1 ensures that each agent produces detailed reasoning in a MAS configuration, across a large number of data points, ensuring scalability.

\subsection{Data Analysis Methods for RQ1}
\label{sec:method:analysis:quant}

\subsubsection{\textbf{Data Preparation: Parsing Reasoning Traces}}
\label{sec:method:analysis:parsing}

DeepSeek-R1 was employed to generate dual-agent discussions, but the raw outputs required systematic parsing prior to analysis. Each discussion contained either two turns (Round 1 independent coding by Agent 1 and Agent 2) or four turns (with the additional Round 2 critique and potential revision). Initially, we parsed the agent's text into three components: (i) the section captured within \verb|<think>...</think>|, which, for lack of a better word, is referred to as reasoning traces; (ii) a short natural-language explanation; and (iii) the final coding decision formatted as a Python-style dictionary. Another parsing step transformed each component into structured fields, enabling subsequent embedding and similarity analyses.
Because the LLM agents did not always produce consistent formatting, these parsing procedures required multiple rounds of debugging and validation. Once we confirmed that all parsing steps were robust, we generated embeddings for the reasoning components.
Within the reasoning component, each agent systematically evaluated all codes specified in the codebook, providing justifications for the presence or absence of each label in relation to the target text. In each round, the agents' attempt at reasoning for each code was then compared to quantify semantic alignment. These reasoning texts were embedded using BERT, averaging token embeddings into a single vector representing each trace's semantic content. When exceeding BERT's 512-token input limit, sequences were truncated.

\subsubsection{\textbf{Data Preparation: Embedding Reasoning Traces and Measuring Similarity}}
\label{sec:method:analysis:baseline}
To quantify semantic alignment between agents, we computed Cosine Similarity (CS) on their reasoning traces. CS was calculated pairwise between Agent 1 and Agent 2 across all segments, yielding a continuous measure of reasoning similarity directly comparable to categorical code agreement.

\subsection{Data Analysis: Within- and Between-Code Agreement}
\label{sec:method:analysis:within-between}
To address RQ1, we investigated whether the CS of agent reasoning traces serves as a valid indicator of code-level alignment. We decomposed agreement into four categories: (1) \textit{within-align}, where agents selected the same label and provided highly similar rationales; (2) \textit{within-misalign}, where agents converged on the same label but justified it with divergent reasoning; (3) \textit{between-align}, where agents produced different labels despite generating semantically similar rationales; and (4) \textit{between-misalign}, where agents disagreed on both the label and the rationale. This framework distinguished cases of strong consensus, surface-level agreement, fuzzy boundaries, and genuine divergence.

For statistical validation, we conducted two complementary analyses. First, we correlated CS with a binary code agreement variable (1 = same label, 0 = different labels) to provide an overall validity coefficient. Second, we used inferential tests, including Welch's $t$-test and effect size estimation with Cohen's $d$, to assess whether reasoning similarity was significantly higher in within-code than between-code cases.

\subsection{Qualitative Analysis Methods for RQ2}
\label{sec:analysis:qual}
To complement the quantitative validation of CS, we conducted a qualitative analysis of agent reasoning traces stemming from their disagreements (RQ2), to examine whether reasoning disagreements surfaced latent sources of uncertainty and ambiguity. We focused on two categories of disagreement: \textit{within-code misalignment} (i.e., same label but different rationales) and \textit{between-code alignment} (i.e., different labels but similar rationales) which reveal ambiguity that could be relevant to pedagogy. For the within-code analysis, we randomly sampled 15 cases from each of 8 representative categories described in \ref{sec:method:codebook}, yielding a total of 120 cases with mid-range CS values (0.55--0.78). After initial review, the human coders excluded \textit{Understanding/Engagement} and \textit{Technical and Logistical Issues}, as these categories did not surface pedagogical points and were therefore dropped from further analysis. For the between-code analysis, we examined 45 cases with high CS values (0.95--0.99), which highlighted fuzzy or overlapping codebook boundaries.

Two human coders independently reviewed these cases under double-blind conditions, recording observations in spreadsheets and Word documents. Their analysis was guided by the central question: \textit{``Do the agents' reasoning traces surface pedagogically-relevant points?"} Coders examined (a) how agents justified their coding decisions, and (b) whether reasoning revealed ambiguity or overlap in codebook categories. After an independent review, coders met to reconcile discrepancies and reached consensus by organizing findings into shared themes (e.g., fuzzy boundaries, ambiguity within codes).

\section{Results}
\subsection{\textbf{RQ1: Validity and Robustness of Cosine Similarity}}
\label{sec:method:analysis:baseline-validation and robustness}
When evaluating whether cosine similarity reflects code agreement, our analysis of 9,746 agent--agent pairs revealed a moderate positive correlation ($\rho = .54,\; 95\%\ \text{CI}\;= [.52,\; .55],\; p < .001$) between reasoning similarity and label agreement. This confirms that label selections typically align with reasoning similarity, though exceptions appear in cases of within-code misalignment (same label, differing rationale) and between-code alignment (different labels, similar rationale).

\begin{table*}[ht]
\centering
\caption{Cosine similarity distribution across levels of agreement and alignment, with 95\% confidence intervals (CI).}
\label{tab:agreement-distribution}
\begin{tabular}{lccc}
\toprule
\textbf{Level} & \textbf{Mean cosine (95\% CI)} & \textbf{N} & \textbf{\% Alignment (95\% CI)} \\
\midrule
Within-align code     & 0.965 [0.964, 0.966] & 4598 & 47.0\% [46.0, 48.0] \\
Between-misalign code & 0.863 [0.861, 0.865] & 2680 & 28.0\% [27.0, 29.0] \\
Within-misalign code  & 0.916 [0.914, 0.918] & 2193 & 23.0\% [22.2, 23.9] \\
Between-align code    & 0.954 [0.951, 0.957] &  275 &  3.0\% [2.7, 3.4] \\
\bottomrule
\end{tabular}
\end{table*}

Nearly half of all cases (47\%) fell under \textit{within-align}, i.e., same label supported by highly similar reasoning (mean cosine = .97). Over a quarter (28\%) were \textit{between-misalign}, where agents disagreed both on the label and the rationale $M_{\text{cosine}} = 0.86$. Another 23\% were \textit{within-misalign}, with identical labels but divergent rationale (mean cosine = .92), suggesting that single codes often encompass diverse instructional sub-functions. Only 3\% fell into \textit{between-align}, where labels differ despite closely matched reasoning (mean cosine = .95), indicating the few instances of fuzzy boundaries in the codebook.

Statistical testing confirmed that agreement pairs ($M = 0.957,\; SD = 0.025$) have significantly higher reasoning similarity than disagreement pairs ($M = 0.904,\; SD = 0.058$), with Welch's $t(9746) = 60.33$, $p < .001$, and a large effect size $d = 1.16$, validating cosine similarity as a robust indicator of alignment.

\begin{figure*}[t]
    \centering
    \begin{subfigure}[b]{0.49\textwidth}
        \centering
        \includegraphics[width=\linewidth]{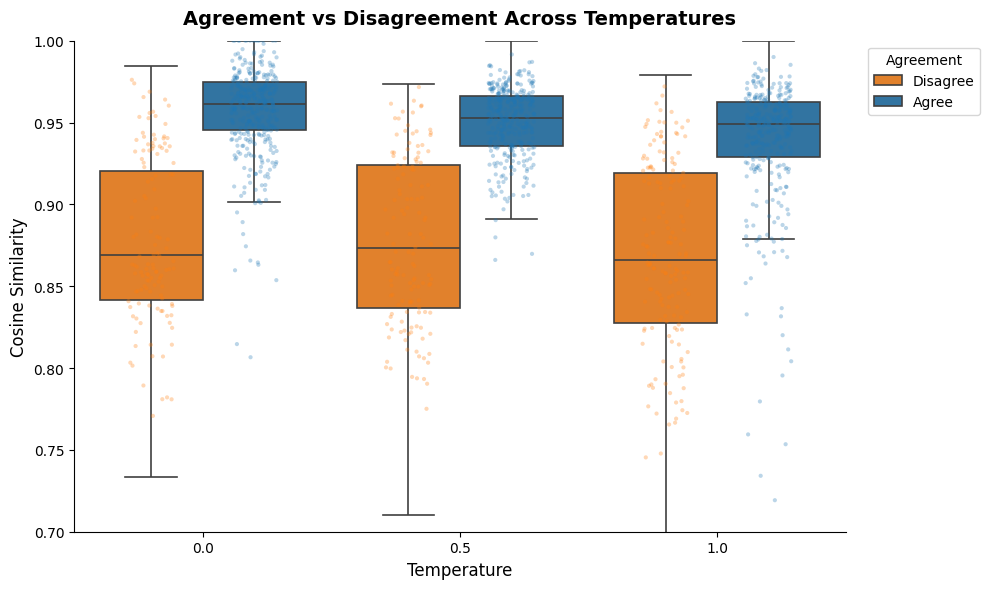}
        \caption{Cosine similarity across temperature settings}
        \Description{A box plot showing cosine similarity distributions for agreement and disagreement pairs across three temperature settings (0, 0.5, 1.0). In all three conditions, agreement pairs cluster tightly around a high similarity score of 0.97, while disagreement pairs are lower and more spread out around 0.90.}
        \label{fig:cosine_temp}
    \end{subfigure}
    \hfill
    \begin{subfigure}[b]{0.49\textwidth}
        \centering
        \includegraphics[width=\linewidth]{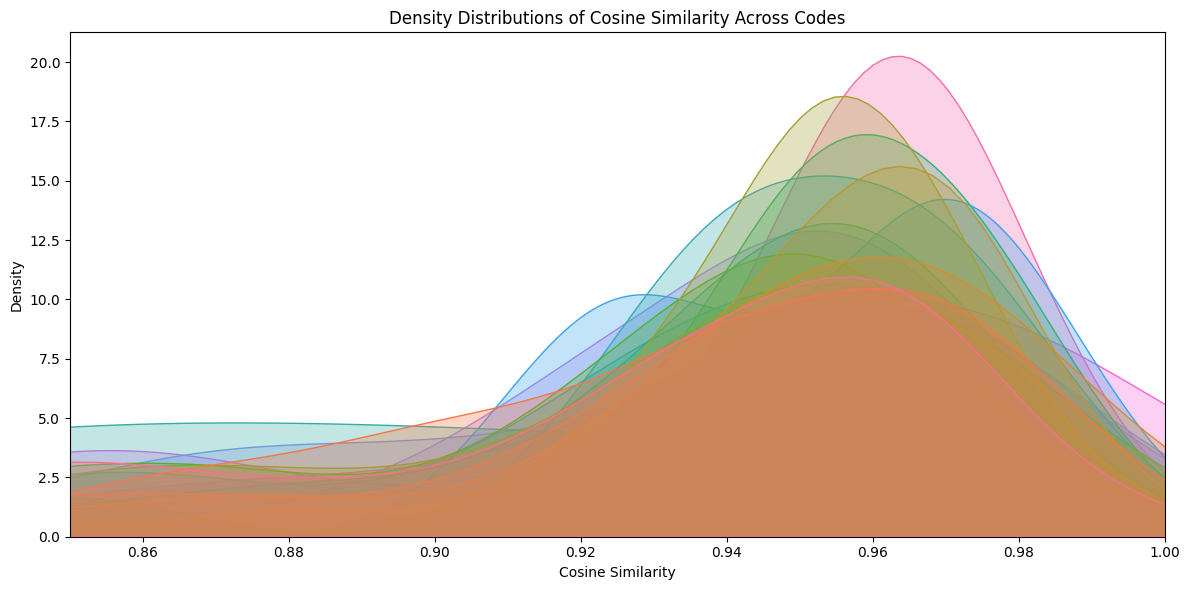}
        \caption{Density distribution by code category}
        \Description{Density plots showing the distribution of cosine similarity scores for different code categories. High-consensus codes like Greeting and Encouragement show sharp peaks near 1.0, while more ambiguous codes like Aligning to Prior Knowledge show flatter, wider distributions indicating lower agreement.}
        \label{fig:density_dist}
    \end{subfigure}
    \caption{Validation of cosine similarity as a metric for reasoning alignment.
    (a) The metric is robust, as the distinction between agreement and disagreement pairs remains stable across temperature settings.
    (b) The metric is sensitive, with distributions revealing interpretive ambiguity across different code categories.}
    \label{fig:combined_similarity}
\end{figure*}

Figure \ref{fig:cosine_temp} shows that these similarity patterns hold consistently across temperature settings (0, 0.5, 1). Agreement pairs cluster tightly around high similarity ($\approx 0.95 \text{--} 0.97$), while disagreement pairs remain distinctly lower with more dispersion ($\approx 0.87 \text{--} 0.92$). Median values stay stable across temperatures, indicating that stochastic sampling variation does not compromise the metric's discrimination.

Figure \ref{fig:density_dist} illustrates how cosine similarity values are distributed across code categories. As shown in Figure \ref{fig:density_dist}, all CS values fall above zero, reflecting the shared semantic grounding of agent reasoning. Sharp peaks in the $0.95\text{--}1.0$ range indicate tight and unambiguous coding, where agents generated nearly identical rationales. To contextualize these patterns, we compare cosine similarity (CS) distributions with findings from prior empirical work reporting human inter-coder reliability ($\kappa$) using the same dataset and codebook \cite{barany2024chatgpt}. For instance, codes such as \textit{Greeting} ($\kappa$ = 0.85) and \textit{Encouragement} ($\kappa$ = 0.80) demonstrated high human inter-coder reliability and correspondingly exhibit tightly concentrated cosine similarity distributions. By contrast, flatter CS distributions correspond to codes with lower $\kappa$, reflecting ambiguity or multiple interpretations. For example, Aligning to Prior Knowledge ($\kappa$ = 0.66) and Checking Understanding/Engagement ($\kappa$ = 0.60), Instruction ($\kappa$ = 0.66), and Guiding Feedback ($\kappa$ = 0.66) as reported by \cite{barany2024chatgpt}, exhibited only moderate consensus among human coders, which is mirrored by the wider spread of their cosine similarity distributions.

Taken together, these quantitative results demonstrate that cosine similarity effectively captures code-level alignment while highlighting interpretive fuzziness, supporting its use as a practical metric for identifying reliable and ambiguous cases in LLM-assisted qualitative coding.

\subsection{RQ2: Qualitative Patterns in Reasoning Traces}

\subsubsection{Within-code Misalignment}
Despite converging on the same label, agents offered different rationales, surfacing heterogeneity within codes or applying different coding approaches. For instance, one agent would follow the prompt procedure and systematically mention all codes in its reasoning trace, while the other agent would directly mention the appropriate codes. Other differences related to the provided source for the argument - while in some cases, agents argued with the codebook definitions, in other cases, agents would relate to the data point. For reasoning traces that diverged on labels of \textbf{Guiding Feedback (GF)}, the justifications ranged from confirmation (e.g., \textit{``yes/that's correct"}), to correction (e.g., nudging in the case of errors), to clarification (e.g., step-by-step explanation). All were coded as GF but grounded in these different instructional functions. For \textbf{Instruction}, rationales were divided into task framers (\textit{``to solve for n"}), procedural steps (\textit{``bring down x"}), modeling (\textit{``let me write it up"}), and directives (\textit{``please write this down"}). Although all utterances were instructional, agents emphasized different forms of expression. For \textbf{Greeting}, agents justified the same label through varied social functions, including check-ins (\textit{``How you doing?"}), apologies, colloquial openers (\textit{``What's good?"}), and closings (\textit{``Have a good day"}). For \textbf{Aligning to Prior Knowledge (ATP}, justifications spanned explicit recall (\textit{``remember\ldots"}), session continuity (\textit{``last time we talked about\ldots"}), and subject anchors (\textit{``this is algebra"}), with some cases overlapping with questioning. For \textbf{Encouragement}, rationales differed between affirmations (\textit{``Yes you can"}), enthusiastic exclamations (\textit{``woo"}), and supportive wishes (``Good luck"). Finally, for \textbf{Time Management}, agents justified codes using pacing reminders (``that should go faster") and session boundaries (``we are kind of at the end"). , and elapsed/remaining time markers (\textit{``five gone," ``class is halfway over"}). In summary, reasoning traces contained a variety of different rationales for the same code labels.

\subsubsection{Between-Code Alignment}
We investigated disagreement in which agents picked different labels despite a high cosine similarity of justifications. Coder 2 identified 9 out of 22 cases where a reasoning trace was present for the label the other agent had applied, but with qualifiers of uncertainty like ``could be/fit" or ``might relate to". In 12 data points, a rationale \textit{against} the label was present, e.g. \textit{``This could be instruction, but wait it's the student speaking"}. The utterance \textit{``Yeah, so use what you learned from the first one and do the second one"} received two labels from agent one (\textbf{ATP} and \textbf{Instruction}), while agent 2 only coded \textbf{ATP}, expressing doubt about assigning two labels (\textit{``It does give a direction, but it also refers back to prior knowledge"}). Short phrases were especially contested. Furthermore, for \textit{``All right, we got two"}, one agent selected \textbf{Encouragement} (a positive acknowledgement) while the other assigned it to \textbf{Instruction} (directing attention to progress in the task). These patterns are indicative of the different perspectives from which single data points can be considered. Uncertainty in turn might point towards overlapping concepts in the codebook or a too narrow context window available to the MAS system.

\section{Discussion and Implications}

Together, these quantitative and qualitative analyses show that cosine similarity robustly identifies alignment and divergence in multi-agent reasoning, while qualitative findings reveals important pedagogical nuances and codebook boundaries. These findings support disagreement as a rich analytic signal for advancing learning analytics methodologies.

The field of learning analytics is rapidly integrating large language models (LLMs) as powerful analytical tools. There is a growing consensus that these tools should augment rather than replace human expertise when interpreting complex qualitative data \cite{barany2024chatgpt, simon2025comparing}. However, systematic methods to scaffold meaningful human--AI collaborative workflows remain underdeveloped. Our study addresses this critical gap by conceptualizing disagreements among agents in a multi-agent LLM system not as mere noise or errors but as rich analytic signals that can meaningfully guide human review.

By analyzing reasoning traces within disagreements --- detected via semantic distance metrics such as cosine similarity --- we provide a rigorous framework to identify and prioritize segments needing human interpretive attention. This approach optimizes the use of limited human coding resources by highlighting codebook ambiguity. Our results show that reasoning-trace embeddings derived from LLM agents serve not only as analytic tools but also hold practical value for generating confidence scores that predict when new data points may require human intervention. This aligns with recent applications of embedding methods to curriculum mapping and human-articulated educational decisions \cite{pardos2019data,xu2023convincing}, suggesting a promising avenue for embedding-driven human--AI partnerships. Furthermore, beyond validating cosine similarity as a robust quantitative measure, our qualitative analysis shows that agent disagreements surface important instructional subtleties and fuzzy or overlapping codebook boundaries---nuances that human coders may overlook when working alone \cite{barany2024chatgpt}. These findings offer actionable insights for codebook refinement and instructional analysis, emphasizing the value of disagreement as a novel analytic signal in learning analytics.

Taken together, our analyses indicate that disagreement in LLM agents' reasoning traces can be treated as a valuable analytic modality rather than mere noise. We argue that these traces become valid and meaningful when interpreted through structured human oversight, not as standalone evidence. Accordingly, combining LLM reasoning agents with human oversight can flag ambiguous cases, support codebook refinement, and strengthen the validity of qualitative findings in learning analytics.

\subsection{Limitations and future work}

Several limitations suggest directions for future research. First, a limitation of our study is that we evaluated the method on a single learner--tutor dialogue dataset. The methodology we advance has broader applicability beyond learner--tutor dialogue datasets and may extend to multimodal educational data streams, including log files, think-aloud protocols, and discussion forums \cite{blikstein2013multimodal}. Incorporating such data into hybrid coding workflows could enhance accuracy, reliability, and trustworthiness, positioning LLM-based analytical tools as supportive partners in authentic educational research contexts \cite{yan2024human,zambrano2023ncoder}. Second, our qualitative investigation sampled only a modest number of within-code misalignments and between-code alignment cases. Broader qualitative analyses are necessary to capture additional instructional patterns and codebook refinements. Third, we did not systematically compare each agent's rationale against codebook definitions, which could illuminate where agent consensus or divergence collectively aligns or deviates from codebook intent --- critical for codebook validation. Fourth, our study focused on a single dataset and coding schema, limiting generalizability. Future work should evaluate the robustness of cosine similarity-based disagreement analysis across diverse datasets, domains, and codebooks. Fifth, the current MAS processes one dialogue segment at a time, which risks ambiguity from limited conversational context and may contribute to divergent codings. Future models incorporating multi-turn or session-level context could reduce this effect. Finally, integrating disagreement signals into practical human--AI workflows remains subject of future research. Designing user-friendly interfaces and annotation tools that help coders interpret and act on these signals is a promising area for further development to enhance coding reliability, efficiency, and pedagogical value in educational research.

\section{Conclusion}

As researchers increasingly turn to automated methods to code and annotate data, LLM reasoning traces and explanations pose a promising and novel source of data for learning analytics.
This study demonstrates that cosine similarity applied to these reasoning traces within multi-agent systems effectively detects meaningful disagreements during qualitative coding, quantifying the degree of conceptual ambiguity in constructs. Treating these traces as data enriches learning analytics by providing quantitative process metrics that complement traditional agreement metrics (e.g., inter-rater reliability) during human coding.

Qualitative analyses reveal that LLM-to-LLM disagreement data illuminate codebook ambiguities and can underpin improved coding workflows. By combining quantitative and qualitative approaches, our methodology supports a human--AI collaborative paradigm that can enhance methodological rigor and interpretive depth in educational research. Our findings highlight disagreement as a valuable new analytic signal in learning analytics, with broad potential to advance qualitative coding, instructional analysis, codebook development, and ultimately, the design of automated coding tools embedded in human-centered workflows.

\begin{acks}
We thank the Penn Center for Learning Analytics (especially Amanda Barany, Xiner Liu, and Nidhi Nasiar) and Saga Education for support in data and codebook curation. We also acknowledge Research Computing at Arizona State University for providing computing resources that have contributed to the research results reported in this manuscript.
\end{acks}

\bibliographystyle{ACM-Reference-Format}
\bibliography{main}

\end{document}